\documentclass[letterpaper, 10 pt, conference]{ieeeconf}  

\IEEEoverridecommandlockouts                              

\usepackage{graphics} 
\usepackage{epsfig} 
\usepackage{times} 
\usepackage{amsmath} 
\usepackage{amssymb}  
\usepackage{algorithm}
\usepackage[noend]{algpseudocode}
\usepackage{xcolor}
\usepackage{balance}

\title{\LARGE \bf
Deep Reinforcement Learning to Acquire Navigation Skills for Wheel-Legged Robots in Complex Environments
}

\author{Xi Chen, Ali Ghadirzadeh, John Folkesson and Patric Jensfelt}

\begin{document}

\maketitle
\thispagestyle{empty}
\pagestyle{empty}

\begin{abstract}
Mobile robot navigation in complex and dynamic environments is a challenging but important problem. 
Reinforcement learning approaches fail to solve these tasks efficiently due to reward sparsities, temporal complexities and high-dimensionality of sensorimotor spaces which are inherent in such problems.
We present a novel approach to train action policies to acquire navigation skills for wheel-legged robots using deep reinforcement learning. 
The policy maps height-map image observations to motor commands to navigate to a target position while avoiding obstacles. We propose to acquire the multifaceted navigation skill by learning and exploiting a number of manageable navigation behaviors. We also introduce a domain randomization technique to improve the versatility of the training samples. 
We demonstrate experimentally a significant improvement in terms of data-efficiency, success rate, robustness against irrelevant sensory data, and also the quality of the maneuver skills.  





\end{abstract}


\section{INTRODUCTION}
Deep Reinforcement Learning (RL) has enabled training of highly flexible and versatile deep neural networks to obtain action-selection policies for complex problems. The complexity generally arises from (1) complicated dynamics and the way actions at different time-steps affect the long-term outcomes, (2) high-dimensionality of the action space which makes it impossible to enumerate actions using classical approaches, and (3) high-dimensionality and redundancies of the sensory observations. In such cases, deep RL holds the promise of finding solutions, sometimes, demonstrating superior performances compared to hand-crafted solutions or even compared to a human-expert himself performing the task \cite{silver2017mastering,mnih2015human}. 
However, the state-of-the-art methods in deep RL are not generally applicable in other problem domains. The methods suffer from issues such as (1) reward sparsity, i.e., low-probable reward outcome while randomly exploring consequences of actions, (2) temporal credit assignment which refers to the problem of crediting action-decisions made over a period of time given a reward/punishment outcome, (3) data inefficiency, i.e., requiring huge amount of training samples to obtain a policy, and (4) difficulties of learning a task-relevant representation of input data which is critical to acquiring a generalizable action-selection policy. 

In this paper, we propose an RL approach to solve a complex mobile robot navigation problem. We train a deep neural network policy to control a mobile robot with high-dimensional action spaces enabling complicated maneuvers, e.g, slimming the body to pass narrow corridors or lifting to cross over obstacles.
The trained policy directly processes height-map images to generate appropriate action commands. The contribution of this paper is to introduce a method which:

\begin{enumerate}
\item improves policy training by splitting the complicated navigation task into a number of manageable navigation behaviors,
\item proposes a domain randomization technique, guiding policy training to attend important cues of the input observations. We experimentally demonstrate the suitability of the trained policy, e.g., to attend to the obstacles in front of the robot, but not the ones it has already passed. 
\item demonstrates an improvement of 20\%  success rate compared to the state-of-the-art RL methods applied to a challenging mobile robot navigation problem. 
\end{enumerate}

The structure of the paper is as follows: In the next section, we outline related work (Sec.~\ref{sec:relatedwork}), followed by introducing required background (Sec.~\ref{sec:preliminaries}). We introduce our method in Sec.~\ref{sec:methods}, and in Sec.~\ref{sec:experiments}, we present our experimental results. The conclusions and the future work are presented in Sec.~\ref{sec:conclusions}.

\section{RELATED WORK}
\label{sec:relatedwork}
In this section, we introduce recent studies that are mostly related to our work. We first introduce a number of the state-of-the-art deep RL methods, followed by an overview of recent deep RL solutions for mobile robot navigation problems.
Also, we provide a short overview of domain randomization approaches that are used recently to improve RL policy training. 
\subsection{Deep reinforcement learning}
Deep RL approaches to train neural network robot policies can be categorized as, (1) end-to-end policy training, (2) concatenating separately trained neural network blocks, and (3) guided policy search. 
End-to-end policy training methods train deep neural network architectures directly with minimum task-dependent engineering efforts. These methods have been successfully applied to different complex tasks, such as playing Atari games \cite{mnih2015human,mnih2016asynchronous}, and a number of 3D simulated physics tasks, \cite{schulman2015trust,schulman2017proximal,lillicrap2015continuous}. 
However, these approaches are very data-inefficient and may not be directly applicable to real robotic problems. 
The second approaches, generally train perception and motor control layers of a deep policy network separately. These layers are mostly trained by learning a low-dimensional representation of the data using auto-encoder structures. These approaches have been applied to solve complex visuomotor tasks, e.g., \cite{finn2016deep,ghadirzadeh2017deep}. 
A major limitation of these approaches is that the structure of the network, data-representations, and training individual blocks require extra engineering efforts. 
Guided policy search \cite{levine2016end}
trains a deep policy network, end-to-end, efficiently by converting the policy search problem into  supervised learning and trajectory optimization problems. 
However, the limitation of this approach is that it requires access to the true state of the system during the training phase to solve the trajectory optimization part.

Our proposed solution belongs to the first category of the approaches. We improve sample efficiency by splitting the problem into simpler sub-tasks and also by exploiting domain randomization techniques to enhance the versatility of the training samples. 

\subsection{Deep RL solutions to mobile robot navigation}
Navigation learning for a mobile robot with high degrees of freedom, such as a wheel-legged robot, requires learning motor controls which stabilize robot motions without falling and also move the robot to a given target while avoiding collisions with obstacles. 
Previous studies addresses these problems separately.
\cite{schulman2015trust,lillicrap2015continuous,schulman2015high,heess2015learning} demonstrate stable locomotion skills on flat surfaces with no obstacles. 
\cite{peng2016terrain} and \cite{heess2017emergence} consider terrain-adaptive motions which enable the robot to cope with more diverse and challenging sets of terrains and obstacles.
However, these methods do not take into account a target position. 
\cite{tai2017virtual} addresses navigation to a target position by training a policy, end-to-end.
However, the method is only validated on a robotic platform with low degrees of freedom. 
\cite{heess2016learning} and \cite{peng2017deeploco} proposed methods to combine the locomotion and motion planning with hierarchical structures with a two-step procedure: (1) training low-level controllers, (2) acquiring a high-level planner given the trained low-level controllers. 


We focus on end-to-end training of a deep architecture to learn locomotion skills and also reaching to different target positions which requires long-term planning.
Our method is validated on challenging configurations with random start and target positions. 



\subsection{Domain Randomization}
Recently, domain randomization techniques are exploited  to transfer action policies trained in simulation to the real world. 
These approaches randomize different aspects of the tasks, such as dynamics, \cite{peng2017sim}, or visual sensory observations, 
\cite{ghadirzadeh2017deep,tobin2017domain,tobin2017domaintransfer,bousmalis2017using}, 
We apply domain randomization to improve the versatility of the training samples collected from simple environments with few obstacles. The resulting dataset contains more complex configurations with multiple obstacles and challenging pathways. 

\section{PRELIMINARIES}
\label{sec:preliminaries}
In this section, we review related RL algorithms and introduce the notation which is used in the rest of the paper.
We assume a Markov Decision Process (MDP) to represent our system, which is defined by the tuple $(\mathcal{S}, \mathcal{A}, \mathcal{P}, r, p(s_0), \gamma)$, 
where $\mathcal{S}$ is a set of states, 
$\mathcal{A}$ is a set of actions, 
$\mathcal{P}:\mathcal{S}\times\mathcal{A}\times\mathcal{S} \rightarrow [0,1]$ is the transition probability, 
$r:\mathcal{S} \times \mathcal{A} \rightarrow \mathbb{R}$ is the reward function, 
$p(s_0):\mathcal{S}\rightarrow[0, 1]$ is the initial state distribution,
and $\gamma\in[0,1]$ is the discount factor.
The actions are sampled based on a  parameterized stochastic policy $\pi_\theta: \mathcal{S} \times \mathcal{A} \rightarrow[0,1]$, which assigns a probability distribution over the actions conditioned on the state value. 

For each episode, an initial state is drawn from $p(s_0)$. At every time-step $t$, an action $a_t$ is sampled from $\pi_\theta(a_t|s_t)$, a reward $r_t = r(s_t, a_t)$ is given by the environment and the next state $s_{t+1}$ is given according to the transition probability $p(s_{t+1}|s_{t},a_{t})$.
For every state-action pair in the trajectory, the return is defined as the sum of discounted rewards, $R_t = \sum_{t'=t} \gamma^{t'-t} r(s_{t'}, a_{t'})]$.
The goal is to obtain a policy which maximizes the expected return,
$\eta(\pi) = \mathbb{E}[R_t]$,
with respect to all possible trajectories following the introduced sampling procedure.

Actor-critic approaches train a policy in three steps, (1) sampling a number of trajectories using the current policy $\pi_\theta$, (2) estimating a value function representing $V_\pi(s_t)= \mathbb{E}[R_t]$, and (3) updating the policy parameters to increase the likelihood of trajectories with higher returns. 
A common approach, known as the policy gradient method, updates policy parameters to maximize $\eta(\pi)=\mathbb{E}[\log \pi_\theta(a_t|s_t) A_t]$, where $A_t$ is an estimate of the advantage function, found as $A_t = R_t - V(s_t)$. Intuitively, the policy is updated such that state-action pairs with higher advantages become more probable. 

Trust region policy optimization (TRPO) \cite{schulman2015trust} introduced a similar surrogate function,  
\begin{equation*}
\eta_{\pi_\theta}^{TRPO} = \mathbb{E}[\frac{ \pi_\theta(a_t|s_t)}{\pi_{\theta_{old}}(a_t|s_t)}A_t-\beta D_{KL}(\pi_{\theta_{old}} || \pi_\theta)],
\end{equation*}
where, $D_{KL}$ represents Kullback Leibler (KL)-divergence. 
TRPO uses a policy probability ratio, $\psi_\theta = \pi_\theta(a_t|s_t)/\pi_{\theta_{old}}(a_t|s_t)$, instead of $\log \pi_\theta$; furthermore it penalizes deviations from the old policy by an extra weight parameter $\beta$. 
In a more recent work, proximal policy optimization (PPO), \cite{schulman2017proximal} derived an updated version of the TRPO surrogate function by clipping the probability ratio $\psi_\theta$ using a function $\mathcal{Z}(.)$ which clips input values  to the range $[1-\epsilon, 1+\epsilon]$, where $\epsilon$ is a hyper-parameter.
This is equivalent to the TRPO's KL-divergence penalty term but forces the ratio of the current policy and old policy to be close to 1, instead of penalizing the difference of the policies. 
The surrogate function is found as the expected minimum value of the clipped and unclipped probability ratios multiplied by the advantage function, i.e.,
\begin{equation} 
\eta_{\pi_\theta}^{PPO}=\mathbb{E}[ \min(\, \mathcal{Z}(\psi_\theta)A_t \, , \, \psi_\theta A_t\,) ].
\label{eq:ppo}
\end{equation}
In this paper, we optimize the surrogate function defined in Eq.~\ref{eq:ppo} to train action policies.

\section{METHOD}
\label{sec:methods}

In this section, we introduce our method to train an action-selection policy for mobile robot navigation problems using deep reinforcement learning. 
The policy maps a height-map to a number of actions which move the robot to its target location. 
We propose to train several secondary policies, each to acquire a certain behavior, and then combine them to train the primary policy. 
We argue that splitting such a complicated problem into a number of manageable tasks would help the RL agent to overcome difficulties with reward sparsity, credit assignment problem and data inefficiency. 
Furthermore, we introduce a domain randomization technique to efficiently learn to attend task-relevant aspects of the sensory observations without further interactive training using RL.
In the rest of this section, we provide details of our approach regarding 
(1) training the secondary policies, (2) applying domain randomization to improve perception layers of the policy, (3) training the primary policy, and (4) structure of the network. 

\subsection{Policy training to acquire different behaviors}
A behavior is defined as a maneuver strategy to move to a target position while avoiding collisions with an obstacle.
We define the following behaviors for our mobile robot navigation problem:
\begin{enumerate}
\item moving straight to the target position with no obstacle along the path,
\item moving around an obstacle to reach a target,
\item driving over a high obstacle by lifting the body,
\item driving over a short but wide obstacle by lowering the body and stretching the legs out,
\item squeezing the body to pass through narrow corridors.
\end{enumerate}

A simple setup w.r.t. a given behavior is defined as an environment with obstacles in which random action exploration results in higher task success rate realizing the specific behavior. 
For example, for the behavior (3), we make obstacles such that the robot has no other options than moving to the target position while lifting the body to avoid collisions.
We combine the knowledge obtained by the secondary policies to train the general policy. 
Dividing a task into a number of behaviors resembles the way a human acquire a multifaceted skill, e.g., playing tennis. In this example, a trainee practices fore-hand and back-hand hits separately to improve each individual skill, and then combines them in a more realistic play condition.  

Secondary policies are trained using the same method as the primary policy. 
We use an actor-critic framework with a network structure which shares parameters between the policy and the value function. Network parameters, denoted by $\theta$, are found such that the following compound loss function is optimized:
\begin{equation} 
\label{eq:loss}
\mathcal{L}(\theta) = \mathbb{E}[\lambda_1 \eta_{\pi_{\theta}}(\theta)+\lambda_2 \mathcal{L}_{v}(\theta)],
\end{equation}
where, $\mathcal{L}_v$ is the value-function loss defined as $||V_\theta(s_t)-R_t||$, $\eta_{\pi_\theta}$ is as defined in Eq.~\ref{eq:ppo}, and the expectation is found over sampled trajectories as described in Sec.~\ref{sec:preliminaries},

The pseudo-code for training secondary policies is presented in Alg.~\ref{alg:ppo_local}.
For every behavior, we initialize a secondary policy as well as a number of simple environments corresponding to that behavior. The first behavior which is learned is to move straight to the target with no obstacles. 
All other secondary policies are initialized with this trained policy. We train every policy for a number of iterations. A number of trajectories are sampled under the current policy and for every action-state pair in these trajectories, the return, advantage and policy probability ratio are found. Finally, the policy and the value-function are updated with Stochastic Gradient Ascent (SGD) w.r.t. the loss function defined in Eq.~\ref{eq:loss}. The latest trajectories which are found based on the trained policies are stored for later use. 
\begin{algorithm}
  \caption{Training secondary policies} 
  \label{alg:ppo_local}
  \begin{algorithmic}[1]
  	\For{every behavior $b$}
  		\State initialize policy $\pi^b_\theta$.
        \State prepare a number of simple train environments.      
      \For{every environment $E$}
      \For{every iteration}
      	\State sample trajectories given $\pi^b_{\theta}$ and $E$.
        \For{every $(s_{t},a_{t})$ pair in every traj. $i$}        
        \State calculate $R_{t,i}$, $A_{t,i}$, $\psi^b_{\theta_{t,i}}$.        
        \EndFor
        \State $\theta_{old}^b \leftarrow \theta^b$.
        \State update $\theta$ by SGD, w.r.t. Eq.~\ref{eq:loss}.
      \EndFor 
      \State store the latest trajectories.     
      \EndFor       
    \EndFor
  \end{algorithmic}
\end{algorithm}

\subsection{Domain Randomization}
We use domain randomization techniques to help the policy to extract task-relevant aspects of the input observations. 
The policy directly maps an image observation, i.e., the height-map representation of the scene, to the motor commands using a forward pass of the neural network.
Given a limited number of training data gathered by actively interacting with the environment and the flexibility of the network, there is a high risk that the policy attends to task-irrelevant components of the observations, which limits its applicability to unseen test environments.

We randomize non-essential aspects of the task, such as the appearance, the positions and the number of obstacles in the scene to improve generalization capabilities of the final policy. 
This randomization is applied to the simple environments where the secondary policies are trained without affecting the validity of the stored solutions.
In other words, we randomize original environments used by the secondary policies by changing the obstacle configurations such that it would not affect the success of the sequence of motor actions found by the trained secondary policy in the corresponding original environment. 
In this way, we obtain solutions to very complex navigation problems without running the RL agent. The RL agent is data-inefficient, and is not guaranteed to find a solution. 

Every environment used by the secondary policies is randomized to generate a number of new environments. For every new environment, the stored actions corresponding to the original environment are applied sequentially and the action-observation-return tuples are stored. 
In this case, we will have $n_e\times n_\tau \times n_{e_{rnd}}$ new action-observation trajectories, where $n_e$ is the number of original environments, each correspond to $n_\tau$ trajectories, and $n_{e_{rnd}}$ number of randomized environments. 
These new trajectories are used to train the primary policy as explained in the next section.

\subsection{Training the primary policy}
The primary policy is trained with the same method and architecture as the secondary policies but with a different sampling strategy. Trajectories are sampled partially by following the primary policy given a number of new complex environments. The rest of trajectories are directly taken from the batch of domain randomized trajectories without running on the robot. 
At the beginning of the training phase, samples which are drawn from the primary policy mostly fail because of the low-probability of reward events when making random sequential action-decisions. 
This reward sparsity is compensated by the samples drawn from the domain randomized batch which only contains successful trials. 



The pseudo-code for training primary policy is presented in Alg.~\ref{alg:ppo_primary}. The primary policy is initialized either randomly or by any of the secondary policies. 
A number of training environments with complex obstacle configurations are generated. The agent is trained in every environment based on the samples drawn from the real interactions with the environments and also samples from the domain randomized batch. In the latter case, the advantages of the trajectories are re-calculated with the updated value function. At each iteration, the parameters of the policy and the value function are updated using stochastic gradient ascent given the compound training data. 
\begin{algorithm}
  \caption{Training the primary policy}
  \label{alg:ppo_primary}
  \begin{algorithmic}[1]
    \State initialize primary policy $\pi_\theta$.
  	\State prepare a number of complex environments.
    \For{every environment $E$}
      \For{every iteration}      	        
      	\State sample trajectories given $\pi_{\theta}$ and $E$.
        \For{every $(s_{t},a_{t})$ pair in every traj. $i$}
        \State calculate $R_{t,i}$, $A_{t,i}$, $\psi_{\theta_{t,i}}$.
        \EndFor        
      	\State collect tuples $(s_{t},a_{t},R_{t})$ from the batch of traj.
        \For{every $(s_{t,i},a_{t,i},R_{t,i})$ tuples}
        \State calculate $A_{t,i}$ using $V_{\theta}(s_{t,i})$.
        \EndFor        
        \State concatenate training data.
        \State $\theta_{old} \leftarrow \theta$.
        \State update $\theta$ by SGD, w.r.t. Eq.~\ref{eq:loss}.
      \EndFor
      \EndFor   
  \end{algorithmic}
\end{algorithm}

\subsection{Network Architecture}
\begin{figure*}[thpb]
  \centering
  \includegraphics[width=0.9\textwidth]{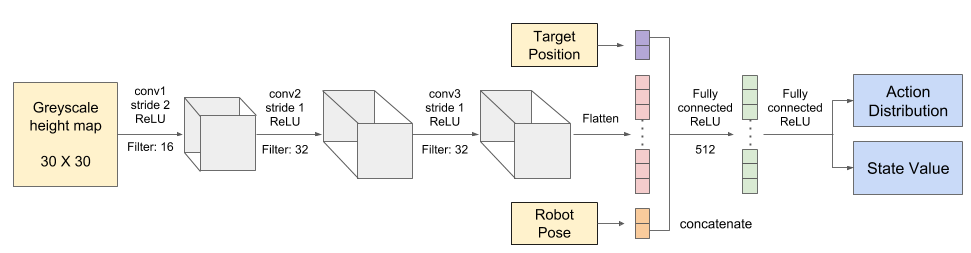}
  \caption{The network architecture used in this paper.}
  \label{fig:network}
\end{figure*}

The network architecture, illustrated in Fig.{\ref{fig:network}}, maps the inputs, consisting of a height-map image observation and the robot and target poses, to a distribution over the motor actions. It also outputs the state value by the value-function sub-network. 
The network consists of three convolutional layers to extract features from the height-map image. The image features are concatenated with the robot configuration and the target position, which are further processed by the fully connected layers to output the action distribution and the state value. 

\section{EXPERIMENT}
\label{sec:experiments}

We want to address the following questions in the experiments: 
\begin{enumerate}
\item can we improve the training efficiency and performance using the batch of domain randomized trajectories? 
\item does the primary policy learn to attend the task-relevant components of the input observation?
\end{enumerate}

To answer these questions, we run a navigation experiment with a reconfigurable wheel-legged robot in simulated environments. We train a baseline model without the randomized trajectory batch, and compare the result to our primary policy trained with the trajectory batch. 
The baseline and the primary policy are trained with the state-of-the-art reinforcement learning algorithm PPO.
Also, we compare the trajectories generated by different observations to investigate if the primary policy learns to attend the task-relevant components of the observation. 

\subsection{Robot and Environment}
\subsubsection{Robot Model}
\begin{figure}[thpb]
  \centering
  \includegraphics[scale=0.38]{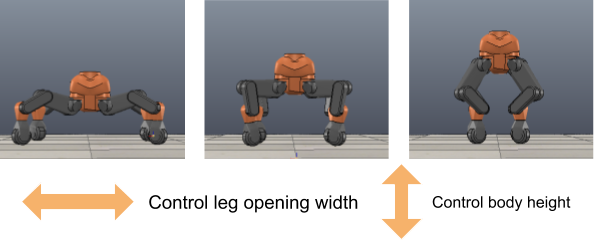}
  \caption{The robot height and the width of leg opening are controlled through separate channels. }       
  \label{fig:leg}
\end{figure}

We assume the wheel-legged robot performs the same action symmetrically to all legs. The height of the body and the width of the leg opening are controlled through the three joints on each leg (Fig.{\ref{fig:leg}}). Together with rotation and translation on the $xy$ plane, the action space of the robot is $5$ dimensional.

We use a discretized value to control each action dimension. For every step, the robot can move $+/- 0.05m$ along $x$ and $y$ axis, rotate $+/-5 degrees$ around the its center axis, can change the body height for $+/-0.02m$, and can stretch legs for $+/-0.02m$.

\subsubsection{Environment Configuration Sampling}
\begin{figure}[thpb]
  \centering
  \includegraphics[width=0.45\textwidth]{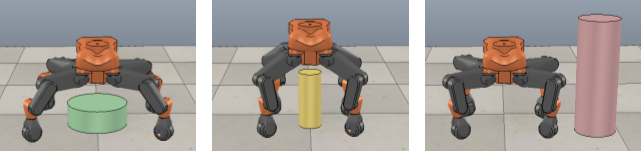}
  \caption{The shape of obstacles and how to interact with them. The robot can drive over the first and the second obstacles by increasing the width of leg openings or lifting the body. The robot has to drive around the third obstacle to avoid collision.}
  \label{fig:obs}
\end{figure}

The environment is simulated by V-REP ({\cite{rohmer2013v}}). 
For each episode, we assign a random orientation to the robot and place the target between $0.5m$ and $1m$ away from the robot along the x-axis of the environment. 
The obstacles in the environment have three different shapes (Fig.{\ref{fig:obs}}) which require the robot to make use of its locomotion skills to either drive around or drive over them. 
The obstacles are placed at the center of a grid-cell with size $0.25m\times0.25m$. The number/shape and position of the obstacles that appear in the scene are randomly assigned.
The height map has the size of $1.6\times1.6m$ centered at the robot. The resolution of the height map is $0.05m/cell$.

In order to give enough exploration for difficult configurations or configurations which have been sampled seldom, we save the configuration which our agent failed to find a solution and assign a probability to re-sample them.
The episode is terminated when the robot drives too far from the target or reach to the maximum number of steps. 
If the robot collides to an obstacle, the environment will be reset to the state before the action is performed.  
The trajectory is considered successful only if the robot reaches the target position.

\subsection{Reward}
The reward function consists of three components: (1) a fixed time penalty for each time step $r_{step}=-0.1$, (2) a progress reward $r_{progress}=d_{t-1}-d_t$, where $d_t$ denotes the distance from the robot to the target at time step $t$, and (3) a fixed penalty $r_{invalid}=-0.1$ when the robot collides with an obstacle, tries to perform an impossible action or moves outside of the boundary. 
\begin{equation}
r(s_{i},a_{i}) = r_{step} + r_{progress} + r_{invalid} 
\end{equation}

The reward function is designed to encourage the agent to reach to the target as soon as possible and reduce invalid actions such as collision. We do not penalize motor cost or specific motion, i.e., move backward or sideways. 

\subsubsection{Domain Randomization}
\begin{figure}[thpb]
  \centering
  \includegraphics[width=0.4\textwidth]{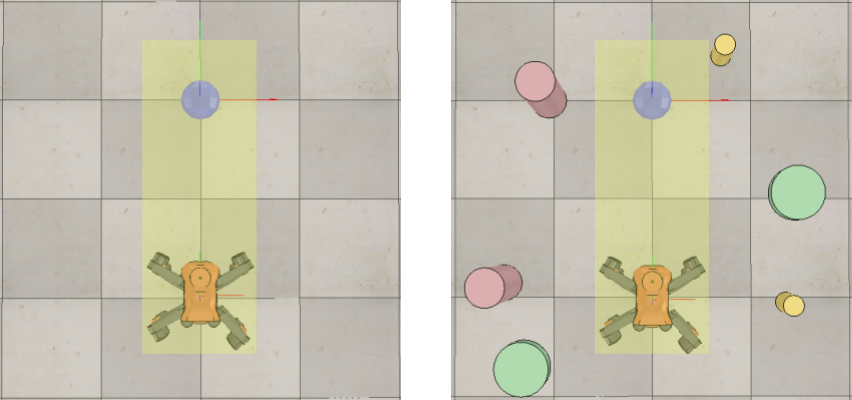}
  \caption{Randomize obstacles without affecting the validity of the solution. The blue point denotes the target position. The yellow square defines the essential area which is affected by a straight line trajectory. The image on the right indicates a randomized environment which does not affect the validity of the straight line solution}
  \label{fig:domain_rend}
\end{figure}

We run each secondary policy and collect trajectories from $10^3$ simple environments. 
For each trajectory generated, we mask the areas of the environment that are affected by the actions in the trajectory. We refer these areas as the essential areas, and randomize a different obstacles configuration in the non-essential areas such that the original trajectory is still valid in the new environment.
We then replay the actions of the original trajectory in sequence and store the new action-observation-return tuples to the trajectory batch. 
For each environment, we repeat the same step for $10^3$ times. Finally we randomize a batch with $5\times10^6$ trajectories. 
Fig.\ref{fig:domain_rend} gives an example of randomizing a new environment from a straight line trajectory.

\subsection{Evaluation}
The secondary policies and the primary policy are trained with the same reward function and network architecture. 
The batch size for training the baseline is $10,000$, for the secondary policy is $3,000$, and for the primary policy is $12,000$, where $10,000$ steps are sampled from the primary policy and $2,000$ steps are randomly sampled from the trajectory batch.  

\subsubsection{Result of Secondary Policy}
\begin{figure}[thpb]
  \centering
  \includegraphics[width=0.47\textwidth]{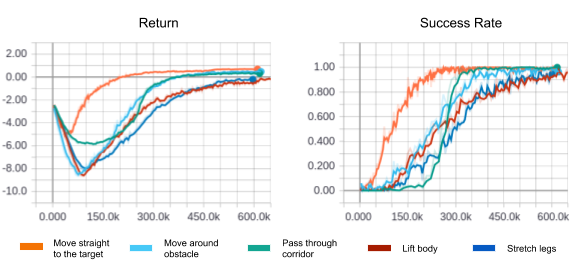}
  \caption{The learning curves of the $5$ secondary policies w.r.t. the average return and success rate. The horizontal axis represents the training steps. }
  \label{fig:secondary_result}
\end{figure}

Fig.~\ref{fig:secondary_result} plots the learning curves for all $5$ secondary policies. 
From the curves, we can see that the agent learns to move straight to the target position with the least amount of training steps since there are no obstacles in the scene.
Moving around one obstacle and passing between two obstacles are similar behaviors. It requires the agent to process the configuration of the obstacles and adjust itself to specific poses and can then move straight to the target position. The learning curves have the similar pattern to the policy of moving straight to the target but need more steps at the beginning.
Learning the behaviors of driving over obstacles by lifting the body and stretching the legs is challenging. The space which allow the robot to move is narrow, which makes the action selection more restricted. The agent needs to learn a rough trajectory first and then gradually optimize it. Compare to other secondary policies, the learning curve of these two policies have smooth and continues increment.  

\subsubsection{Result of Primary Policy}

\begin{figure}[thpb]
  \centering
  \includegraphics[width=0.47\textwidth]{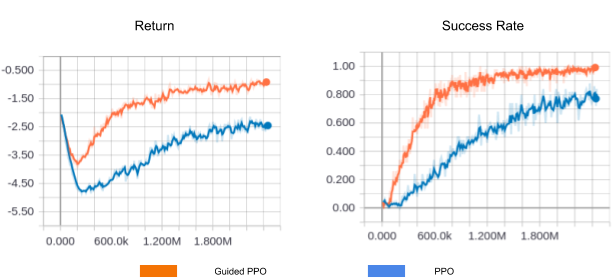}
  \caption{The average return and the success rate of the primary policy. The horizontal axis represents the training steps. }
  \label{fig:result_learn}
\end{figure}

Fig.~\ref{fig:result_learn} plots the learning curves of the primary policy w.r.t. the average return and the success rate. 
The result shows that using the trajectory batch increases the success rate by nearly $20\%$. Especially for situations where the robot needs to lift the body or stretch the legs to drive over obstacles, the baseline model does not generate the appropriate trajectories, which reduces the overall return value.
One of the reasons that may cause the baseline model perform poorly is the temporal credit assignment issue. 
Similar to the problem discussed in \cite{andrychowicz2017hindsight}, the penalty we assign to the collision may hinder exploration, i.e., the agent may learn not to move too close to the obstacles. 
This issue is compensated using the trajectory batch which contains successful actions for similar configurations. 

\begin{figure}[thpb]
  \centering
  \includegraphics[width=0.47\textwidth]{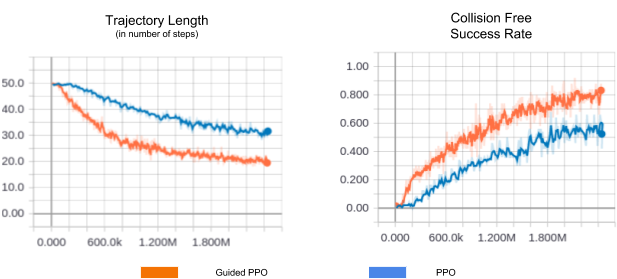}
  \caption{The trajectory length in number of steps and the successful rate for generating a collision free trajectory. The horizontal axis represents the training steps. }
  \label{fig:result_nav}
\end{figure}

It is important that the robot can reach the target position. However, as a navigation task, we need to consider the quality of the trajectory as well. In Fig.\ref{fig:result_nav}, we evaluate the quality of the trajectories w.r.t. the success rate of generating a collision-free trajectory and the length of the trajectory in steps. 
For generating a collision-free trajectory, the model trained with the trajectory batch still reaches a success rate of $80\%$ while the baseline model has only $55\%$.
When we compare the length, the trajectories generated with the batch is $10$ steps shorter than the baseline. 

\subsubsection{Attend task-relevant components of the observation}
In order to verify if the agent learns to attend the task-relevant components of the input observation, we compare the trajectories generated using different observation.
We first sample a configuration and generate a baseline trajectory. We then remove one obstacle from the scene and generate trajectory with one missing obstacle. 
We repeat the same step for all obstacles and compare the trajectories to the baseline to find out which obstacle has more effects on the policy.
We calculate the distance between the two trajectories and use it to measure how much the missing obstacle affects the policy, or how relevant this missing obstacle is to the task.

\begin{figure}[thpb]
  \centering
  \includegraphics[width=0.35\textwidth]{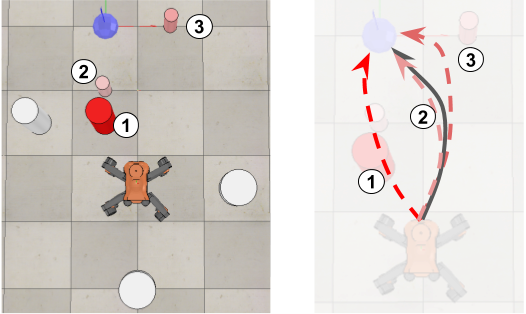}
  \caption{An example of the relevance of each obstacle to the trajectory. In the left image, the relevance of obstacle is color-coded from white to red, where white denotes low relevance and red denotes high relevance. The number on the image denotes the index of the obstacles and the blue point denotes the target position. The trajectories generated are given in the right image. The gray arrow denotes the trajectory generated using all obstacles in the scene. The number near the red trajectories indicates which obstacle is removed from the scene.}
  \label{fig:task_relevant}
\end{figure}

Fig.\ref{fig:task_relevant} gives an example of the relevance of each obstacle. The obstacles colored by white has low relevance, which means the trajectory does not change after removing them. The obstacles colored by red has higher relevance, which can affect or completely change the choice of trajectories. 
From Fig.\ref{fig:task_relevant} we note that the obstacles which close to the current path or block a better path give more impact to the policy.
Our train model is able to attend the task-relevant components from the observation. 

\section{CONCLUSIONS}
\label{sec:conclusion}
In this work, we present a novel approach to learn navigation skills for wheel-legged robot through a number of manageable navigation behaviors. Furthermore, we introduce the domain randomization technique to improve the versatility of the trajectory batch and guide the policy training to attend task-relevant components of the input observation. Our approach overcomes the difficulties with data inefficiency, reward sparsity and temporal credit assignment problem. It outperforms the state-of-the-art approach in terms of both success rate and trajectory quality.

In the future work, we plan to investigate if we can apply our model to more diverse sets of obstacles through domain randomization without further interactive training using RL. Also, we plan to apply our approach to a real robot to perform dynamics tasks like driving on uneven terrain, and to include skills such as walking or climbing, which require higher dimensional actions space

\addtolength{\textheight}{-12cm}   


\bibliographystyle{IEEEtran}
\balance
\bibliography{sections/bib}

\begin{thebibliography}{10}
\providecommand{\url}[1]{#1}
\csname url@samestyle\endcsname
\providecommand{\newblock}{\relax}
\providecommand{\bibinfo}[2]{#2}
\providecommand{\BIBentrySTDinterwordspacing}{\spaceskip=0pt\relax}
\providecommand{\BIBentryALTinterwordstretchfactor}{4}
\providecommand{\BIBentryALTinterwordspacing}{\spaceskip=\fontdimen2\font plus
\BIBentryALTinterwordstretchfactor\fontdimen3\font minus
  \fontdimen4\font\relax}
\providecommand{\BIBforeignlanguage}[2]{{%
\expandafter\ifx\csname l@#1\endcsname\relax
\typeout{** WARNING: IEEEtran.bst: No hyphenation pattern has been}%
\typeout{** loaded for the language `#1'. Using the pattern for}%
\typeout{** the default language instead.}%
\else
\language=\csname l@#1\endcsname
\fi
#2}}
\providecommand{\BIBdecl}{\relax}
\BIBdecl

\bibitem{silver2017mastering}
D.~Silver, J.~Schrittwieser, K.~Simonyan, I.~Antonoglou, A.~Huang, A.~Guez,
  T.~Hubert, L.~Baker, M.~Lai, A.~Bolton \emph{et~al.}, ``Mastering the game of
  go without human knowledge,'' \emph{Nature}, vol. 550, no. 7676, p. 354,
  2017.

\bibitem{mnih2015human}
V.~Mnih, K.~Kavukcuoglu, D.~Silver, A.~A. Rusu, J.~Veness, M.~G. Bellemare,
  A.~Graves, M.~Riedmiller, A.~K. Fidjeland, G.~Ostrovski \emph{et~al.},
  ``Human-level control through deep reinforcement learning,'' \emph{Nature},
  vol. 518, no. 7540, p. 529, 2015.

\bibitem{mnih2016asynchronous}
V.~Mnih, A.~P. Badia, M.~Mirza, A.~Graves, T.~Lillicrap, T.~Harley, D.~Silver,
  and K.~Kavukcuoglu, ``Asynchronous methods for deep reinforcement learning,''
  in \emph{International Conference on Machine Learning}, 2016, pp. 1928--1937.

\bibitem{schulman2015trust}
J.~Schulman, S.~Levine, P.~Abbeel, M.~Jordan, and P.~Moritz, ``Trust region
  policy optimization,'' in \emph{International Conference on Machine
  Learning}, 2015, pp. 1889--1897.

\bibitem{schulman2017proximal}
J.~Schulman, F.~Wolski, P.~Dhariwal, A.~Radford, and O.~Klimov, ``Proximal
  policy optimization algorithms,'' \emph{arXiv preprint arXiv:1707.06347},
  2017.

\bibitem{lillicrap2015continuous}
T.~P. Lillicrap, J.~J. Hunt, A.~Pritzel, N.~Heess, T.~Erez, Y.~Tassa,
  D.~Silver, and D.~Wierstra, ``Continuous control with deep reinforcement
  learning,'' \emph{arXiv preprint arXiv:1509.02971}, 2015.

\bibitem{finn2016deep}
C.~Finn, X.~Y. Tan, Y.~Duan, T.~Darrell, S.~Levine, and P.~Abbeel, ``Deep
  spatial autoencoders for visuomotor learning,'' in \emph{Robotics and
  Automation (ICRA), 2016 IEEE International Conference on}.\hskip 1em plus
  0.5em minus 0.4em\relax IEEE, 2016, pp. 512--519.

\bibitem{ghadirzadeh2017deep}
A.~Ghadirzadeh, A.~Maki, D.~Kragic, and M.~Bj{\"o}rkman, ``Deep predictive
  policy training using reinforcement learning,'' in \emph{Intelligent Robots
  and Systems (IROS), 2017 IEEE/RSJ International Conference on}.\hskip 1em
  plus 0.5em minus 0.4em\relax IEEE, 2017, pp. 2351--2358.

\bibitem{levine2016end}
S.~Levine, C.~Finn, T.~Darrell, and P.~Abbeel, ``End-to-end training of deep
  visuomotor policies,'' \emph{The Journal of Machine Learning Research},
  vol.~17, no.~1, pp. 1334--1373, 2016.

\bibitem{schulman2015high}
J.~Schulman, P.~Moritz, S.~Levine, M.~Jordan, and P.~Abbeel, ``High-dimensional
  continuous control using generalized advantage estimation,'' \emph{arXiv
  preprint arXiv:1506.02438}, 2015.

\bibitem{heess2015learning}
N.~Heess, G.~Wayne, D.~Silver, T.~Lillicrap, T.~Erez, and Y.~Tassa, ``Learning
  continuous control policies by stochastic value gradients,'' in
  \emph{Advances in Neural Information Processing Systems}, 2015, pp.
  2944--2952.

\bibitem{peng2016terrain}
X.~B. Peng, G.~Berseth, and M.~Van~de Panne, ``Terrain-adaptive locomotion
  skills using deep reinforcement learning,'' \emph{ACM Transactions on
  Graphics (TOG)}, vol.~35, no.~4, p.~81, 2016.

\bibitem{heess2017emergence}
N.~Heess, S.~Sriram, J.~Lemmon, J.~Merel, G.~Wayne, Y.~Tassa, T.~Erez, Z.~Wang,
  A.~Eslami, M.~Riedmiller \emph{et~al.}, ``Emergence of locomotion behaviours
  in rich environments,'' \emph{arXiv preprint arXiv:1707.02286}, 2017.

\bibitem{tai2017virtual}
L.~Tai, G.~Paolo, and M.~Liu, ``Virtual-to-real deep reinforcement learning:
  Continuous control of mobile robots for mapless navigation,'' in
  \emph{Intelligent Robots and Systems (IROS), 2017 IEEE/RSJ International
  Conference on}.\hskip 1em plus 0.5em minus 0.4em\relax IEEE, 2017, pp.
  31--36.

\bibitem{heess2016learning}
N.~Heess, G.~Wayne, Y.~Tassa, T.~Lillicrap, M.~Riedmiller, and D.~Silver,
  ``Learning and transfer of modulated locomotor controllers,'' \emph{arXiv
  preprint arXiv:1610.05182}, 2016.

\bibitem{peng2017deeploco}
X.~B. Peng, G.~Berseth, K.~Yin, and M.~Van De~Panne, ``Deeploco: Dynamic
  locomotion skills using hierarchical deep reinforcement learning,'' \emph{ACM
  Transactions on Graphics (TOG)}, vol.~36, no.~4, p.~41, 2017.

\bibitem{peng2017sim}
X.~B. Peng, M.~Andrychowicz, W.~Zaremba, and P.~Abbeel, ``Sim-to-real transfer
  of robotic control with dynamics randomization,'' \emph{arXiv preprint
  arXiv:1710.06537}, 2017.

\bibitem{tobin2017domain}
J.~Tobin, W.~Zaremba, and P.~Abbeel, ``Domain randomization and generative
  models for robotic grasping,'' \emph{arXiv preprint arXiv:1710.06425}, 2017.

\bibitem{tobin2017domaintransfer}
J.~Tobin, R.~Fong, A.~Ray, J.~Schneider, W.~Zaremba, and P.~Abbeel, ``Domain
  randomization for transferring deep neural networks from simulation to the
  real world,'' in \emph{Intelligent Robots and Systems (IROS), 2017 IEEE/RSJ
  International Conference on}.\hskip 1em plus 0.5em minus 0.4em\relax IEEE,
  2017, pp. 23--30.

\bibitem{bousmalis2017using}
K.~Bousmalis, A.~Irpan, P.~Wohlhart, Y.~Bai, M.~Kelcey, M.~Kalakrishnan,
  L.~Downs, J.~Ibarz, P.~Pastor, K.~Konolige \emph{et~al.}, ``Using simulation
  and domain adaptation to improve efficiency of deep robotic grasping,''
  \emph{arXiv preprint arXiv:1709.07857}, 2017.

\bibitem{rohmer2013v}
E.~Rohmer, S.~P. Singh, and M.~Freese, ``V-rep: A versatile and scalable robot
  simulation framework,'' in \emph{Intelligent Robots and Systems (IROS), 2013
  IEEE/RSJ International Conference on}.\hskip 1em plus 0.5em minus 0.4em\relax
  IEEE, 2013, pp. 1321--1326.

\bibitem{andrychowicz2017hindsight}
M.~Andrychowicz, D.~Crow, A.~Ray, J.~Schneider, R.~Fong, P.~Welinder,
  B.~McGrew, J.~Tobin, O.~P. Abbeel, and W.~Zaremba, ``Hindsight experience
  replay,'' in \emph{Advances in Neural Information Processing Systems}, 2017,
  pp. 5055--5065.

\end{thebibliography}

\end{document}